\documentclass{article}

\PassOptionsToPackage{numbers,sort&compress}{natbib}
\usepackage[preprint]{neurips_2026}

\usepackage[utf8]{inputenc}
\usepackage[T1]{fontenc}
\usepackage{hyperref}
\usepackage{url}
\usepackage{booktabs}
\usepackage{amsfonts}
\usepackage{amsmath}
\usepackage{amssymb}
\usepackage{graphicx}
\usepackage{float}
\usepackage{nicefrac}
\usepackage{microtype}
\usepackage{xcolor}
\hypersetup{hypertexnames=false}

\newcommand{\cmark}{\textcolor{red}{\ensuremath{\checkmark}}}
\newcommand{\xmark}{\ensuremath{\times}}

\title{
M\textsuperscript{2}E-UAV: A Benchmark and Analysis for Onboard 
\textcolor{red}{M}otion-on-\textcolor{red}{M}otion 
\textcolor{red}{E}vent-Based Tiny 
\textcolor{red}{UAV} Detection
}

\author{
\normalfont
Weiqi Yan\textsuperscript{1}\thanks{Equal contribution.}
\quad
Lixin Chen\textsuperscript{1}\footnotemark[1]
\quad
Xiangrui Hou\textsuperscript{1}
\quad
Zhipeng Cai\textsuperscript{2}\thanks{Corresponding author.}
\\
Youbiao Wang\textsuperscript{1}
\quad
Yangyang Shi\textsuperscript{2}
\quad
Yu Zang\textsuperscript{1}\footnotemark[2]
\quad
Cheng Wang\textsuperscript{1}
\\[0.5em]
\textsuperscript{1}Fujian Key Laboratory of Urban Intelligent Sensing and Computing, School of Informatics,\\
Xiamen University, Xiamen, China\\
\textsuperscript{2}Meta, Menlo Park, USA\thanks{Meta was not involved in conducting the data collection or experiments discussed in this paper.}\\[0.5em]
\texttt{\{yanweiqi199888, czptc2h, shiyang1983\}@gmail.com}\\
\texttt{\{23020241154341, 23020251154347, 36920241153251\}@stu.xmu.edu.cn}\\
\texttt{zangyu7@126.com, cwang@xmu.edu.cn}
}

\begin{document}

\maketitle

\begin{abstract}
Tiny UAV detection from an onboard event camera is difficult when the observer and target move at the same time. In this motion-on-motion regime, ego-motion activates background edges across buildings, vegetation, and horizon structures, while the UAV may appear as a sparse event cluster. Unlike static- or ground-observer event-based UAV detection, onboard UAV-view detection breaks the clean-background assumption because sensor ego-motion can activate dense background events over the entire field of view. To explore this practical problem, we present M\textsuperscript{2}E-UAV, to the best of our knowledge, the first onboard UAV-view motion-on-motion event-based dataset and benchmark for tiny UAV detection, where both the sensing platform and the target UAV are moving. M\textsuperscript{2}E-UAV provides synchronized event streams and IMU measurements collected from an onboard sensing platform, together with event-level UAV foreground labels derived from temporally propagated 10 Hz bounding-box annotations. The processed benchmark contains 87,223 training samples and 21,395 validation samples across four scene families: sunny building-forest, sunny farm-village, sunset building-forest, and sunset farm-village. We define a train/validation split and an evaluation protocol for comparing representative existing baselines across event-frame, voxel-grid, and point-set representations, with optional IMU input. The benchmark results show that existing baselines remain limited under sparse tiny-target evidence and dense ego-motion-induced background events. Code and benchmark files will be released at \url{https://github.com/Wickyan/M2E-UAV}.
\end{abstract}

\section{Introduction}

Tiny UAV detection is important for safety, inspection, and collision avoidance, but it remains difficult for onboard perception. Conventional RGB or frame cameras can be affected by fast motion, motion blur, illumination changes, and the limited spatial footprint of distant UAVs. Event cameras are attractive for this regime because they respond asynchronously to brightness changes and capture high-temporal-resolution motion cues \citep{gallego2022event}. However, onboard motion-on-motion tiny UAV detection is harder than static-observer detection: the sensing platform and target UAV both move, and ego-motion activates cluttered background edges across buildings, vegetation, and horizon structures.

Existing event-based UAV, flying-object, and tracking datasets are closely related, but most of them are collected from static or ground-based observers. In these settings, the event camera is usually fixed or has limited ego-motion, so moving UAV targets can be detected against relatively clean or weakly activated backgrounds. This assumption breaks down in onboard UAV-view perception. When the event camera itself moves with the carrier UAV, background structures such as buildings, vegetation, terrain boundaries, and horizon edges are continuously activated by sensor ego-motion, producing dense full-field event activity. As a result, the tiny target UAV is no longer the dominant moving source, but only a sparse event cluster embedded in strong ego-motion-induced background events. This changes event-based UAV detection from a relatively clean moving-target detection problem into a new motion-on-motion perception challenge, where both the observer and the target are moving. This gap motivates M$^2$E-UAV as a dedicated benchmark for onboard motion-on-motion event-based tiny UAV detection.

\begin{figure}[t]
  \centering
  \begin{tabular}{@{}cc@{}}
    \includegraphics[width=0.5\linewidth,height=0.16\textheight,keepaspectratio]{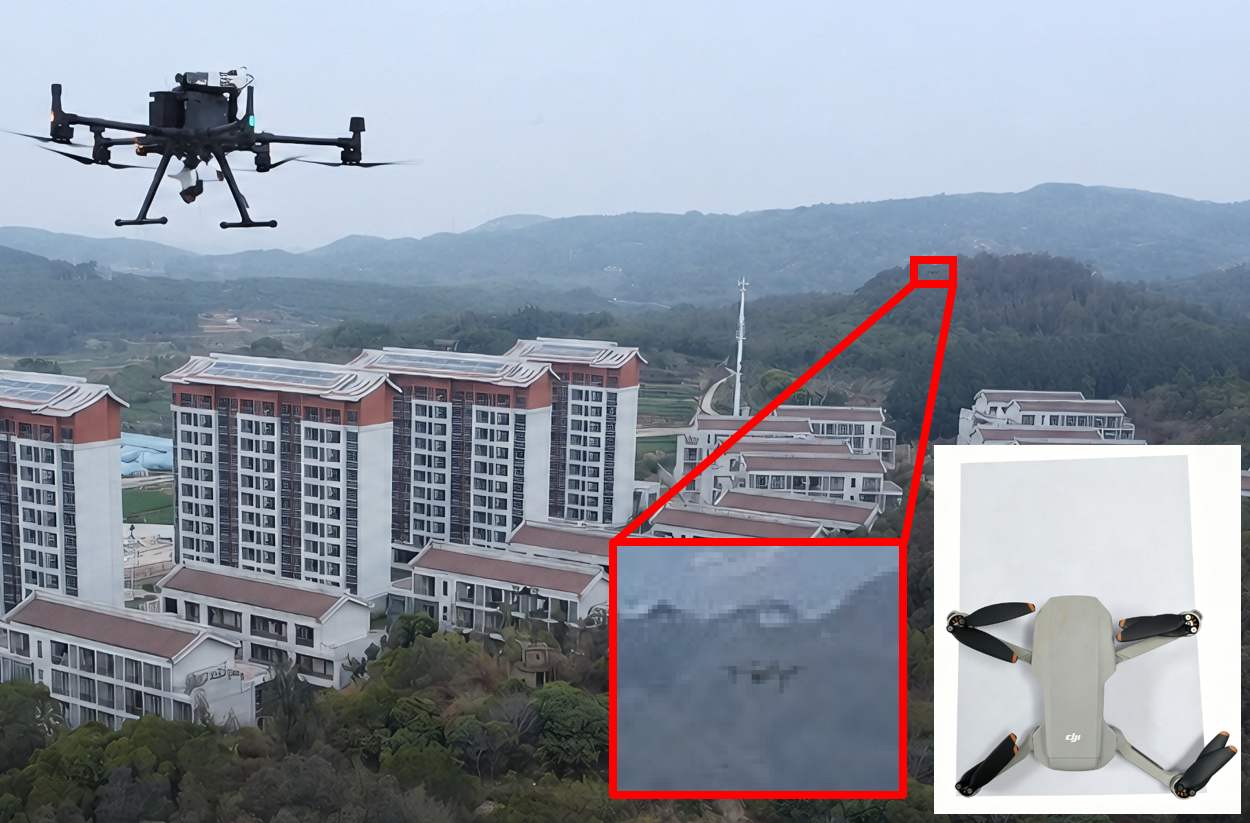} &
    \includegraphics[width=0.5\linewidth,height=0.16\textheight,keepaspectratio]{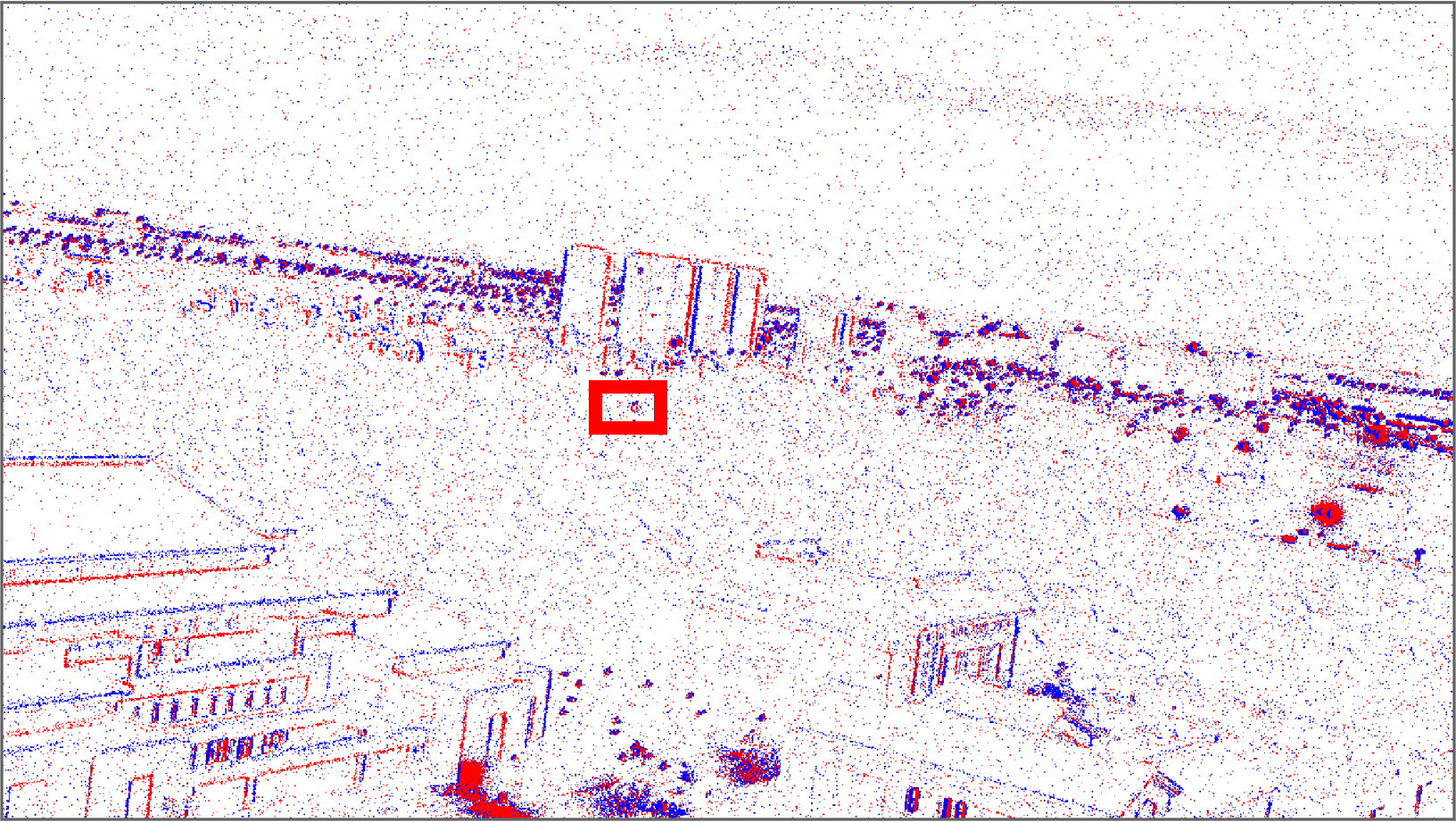}
  \end{tabular}
  \caption{
  \textbf{Teaser of the M\textsuperscript{2}E-UAV onboard motion-on-motion setting.} Left: schematic and real data-collection scene in which a carrier UAV observes a tiny target UAV. Right: actual onboard event-camera data under sensor ego-motion, where background structures induce dense event clutter around sparse target evidence.
  }
  \label{fig:teaser}
\end{figure}

Figure~\ref{fig:teaser} summarizes the proposed setting: the left panel illustrates the real onboard collection scenario in which a carrier UAV observes a tiny target UAV, while the right panel shows actual event-camera observations under sensor ego-motion, where background structures create dense event clutter around sparse target evidence. We construct M\textsuperscript{2}E-UAV as a dataset and benchmark for onboard motion-on-motion event-based tiny UAV detection. M\textsuperscript{2}E-UAV provides 1280$\times$720 event data and synchronized event-IMU packets, where event and IMU are synchronized by an STM32-based trigger synchronization board. The processed benchmark contains 87,223 training samples and 21,395 validation samples across four scene families: sunny building-forest, sunny farm-village, sunset building-forest, and sunset farm-village. 2D UAV bounding boxes are annotated at 10 Hz, sparse boxes are temporally propagated/interpolated, and the resulting supervision is lifted to point-level/event-level UAV foreground labels.

Beyond dataset construction, this paper benchmarks representative existing baselines across event representations, including event-frame detectors, event-tensor methods, and event-point methods, with optional IMU input when supported. These comparisons are intended to characterize the difficulty of M\textsuperscript{2}E-UAV rather than to introduce a new detector.

Our contributions are:
\begin{itemize}
    \item We introduce M\textsuperscript{2}E-UAV, to the best of our knowledge the first event-based dataset and benchmark for onboard motion-on-motion tiny UAV detection, where a moving UAV-mounted event camera detects a moving tiny UAV target.

    \item We provide synchronized 720p event streams and IMU measurements across four illumination-background scene families, together with 10 Hz bounding-box annotations and derived point-level/event-level UAV foreground labels.

    \item We establish a unified benchmark protocol and evaluate representative event-frame, event-tensor, and event-point baselines, showing that sparse tiny-target evidence and dense ego-motion-induced background events remain challenging for current methods.
\end{itemize}

\section{Related Work}

\paragraph{Event-based object detection.}
Event-based object detection has been studied through several input representations. A common direction converts events into frame-like or voxel-like tensors and then applies image-style or video-style detection backbones. Recent event-native detectors further process event streams with recurrent or transformer-based architectures, such as RVT~\citep{gehrig2023rvt} and SAST~\citep{peng2024sast}. These methods show that event tensors can support object detection under high-speed and high-dynamic-range conditions. However, onboard motion-on-motion tiny UAV detection presents a different difficulty: ego-motion can activate dense background edges, while the target UAV may appear as only a sparse event cluster. These limitations highlight the need for a dedicated benchmark for motion-on-motion event-based tiny UAV detection.

\paragraph{Event-based UAV and flying-object benchmarks.}
Several recent datasets study UAVs, drones, flying objects, or event-based tracking. F-UAV-D targets fast UAV detection using dynamic vision sensors and embedded inference~\citep{fuavd_dataset}. NeRDD introduces synchronized RGB-event drone detection data~\citep{nerdd_dataset}, and FRED provides a large RGB-event drone dataset for drone perception~\citep{fred_dataset}. EV-Flying studies in-the-wild recognition of flying objects using event cameras~\citep{evflying_dataset}. EV-UAV focuses on event-based tiny object detection and provides event-level annotations for anti-UAV scenarios~\citep{chen2025evuav}. In addition, event-based tracking benchmarks such as VisEvent~\citep{visevent_dataset}, EventVOT~\citep{eventvot_dataset}, and CRSOT~\citep{crsot_dataset} provide related perspectives on event or RGB-event tracking. Most existing event-based UAV and flying-object resources are collected with static, ground-based, or only weakly moving observers, and the target often acts as the main event source. They therefore do not expose the key difficulty of onboard UAV-view detection: sensor ego-motion continuously activates background structures, so background events can dominate the event stream. M\textsuperscript{2}E-UAV specifically benchmarks this motion-on-motion regime, where a moving UAV-mounted event camera observes a moving tiny UAV target under dense ego-motion-induced background activity.

\paragraph{Point-based event modeling.}
Point-set neural networks provide a natural tool for sparse geometric data. Representative point-cloud segmentation architectures include DGCNN~\citep{wang2019dgcnn}, KPConv~\citep{thomas2019kpconv}, RandLA-Net~\citep{hu2020randlanet}, and COSeg~\citep{an2024coseg}. Although these methods were not originally designed for onboard event-based tiny UAV detection, they provide useful modeling primitives for event packets represented as $[x,y,t,p]$ point sets. In M\textsuperscript{2}E-UAV, the target UAV often appears as a small sparse event structure, making point-set representations an important part of the benchmark.

\paragraph{Event-IMU fusion and motion cues.}
IMU measurements can provide auxiliary motion information for onboard perception, especially when the sensor platform is moving. In M\textsuperscript{2}E-UAV, IMU is included as a synchronized modality so that future methods can study event-only and event-IMU settings under the same data split. This benchmark framing supports motion-aware event-IMU research without tying the dataset to a specific fusion architecture.

\section{Benchmark Setup}

\subsection{Hardware platform and synchronization}
M\textsuperscript{2}E-UAV is built around an onboard sensing platform for motion-on-motion tiny UAV detection. The sensing platform contains an onboard event camera and an IMU mounted on the carrier UAV. Both the event camera and IMU are mounted underneath the carrier UAV and connected to the platform through a vibration-damping plate. To ensure temporal alignment between the asynchronous event stream and inertial measurements, the two sensors are synchronized through a dedicated STM32-based trigger synchronization board. The processed benchmark is therefore organized as synchronized event-IMU packet pairs, which is important for studying onboard motion-on-motion detection where observer motion directly affects the event stream. The benchmark uses event packets as the primary input for tiny UAV detection. Synchronized IMU measurements are provided as optional auxiliary signals, enabling future methods to study onboard ego-motion-aware event perception under the same train/validation split.

\begin{table}[H]
    \centering
    \caption{Hardware configuration of the M\textsuperscript{2}E-UAV data-collection platform.}
    \label{tab:hardware_config}
    \begin{tabular}{ll}
        \toprule
        Component & Specification \\
        \midrule
        Carrier UAV & DJI Matrice 300 RTK \\
        Target UAV & DJI Mini 2, $159 \times 203 \times 56$ mm \\
        Event camera & SilkyEvCam HD, IMX636 CMOS sensor, 8 mm lens \\
        IMU & FDI Systems DETA100D4G \\
        Synchronization & STM32-based trigger synchronization board \\
        \bottomrule
    \end{tabular}
\end{table}

\begin{figure}[H]
  \centering
  % TODO(figure): Replace this schematic with the final benchmark construction diagram.
  \begin{tabular}{@{}l@{}}
    \includegraphics[width=0.9\linewidth,height=1\textheight,keepaspectratio]{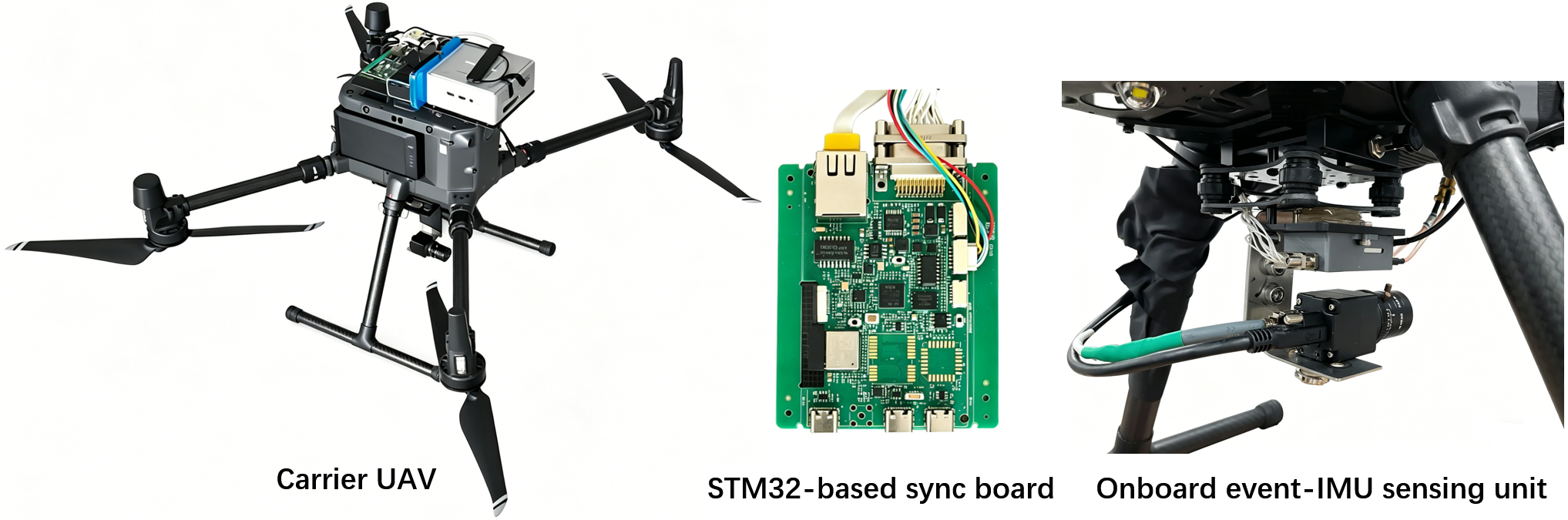} \\
    \includegraphics[width=1\linewidth,height=1\textheight,keepaspectratio]{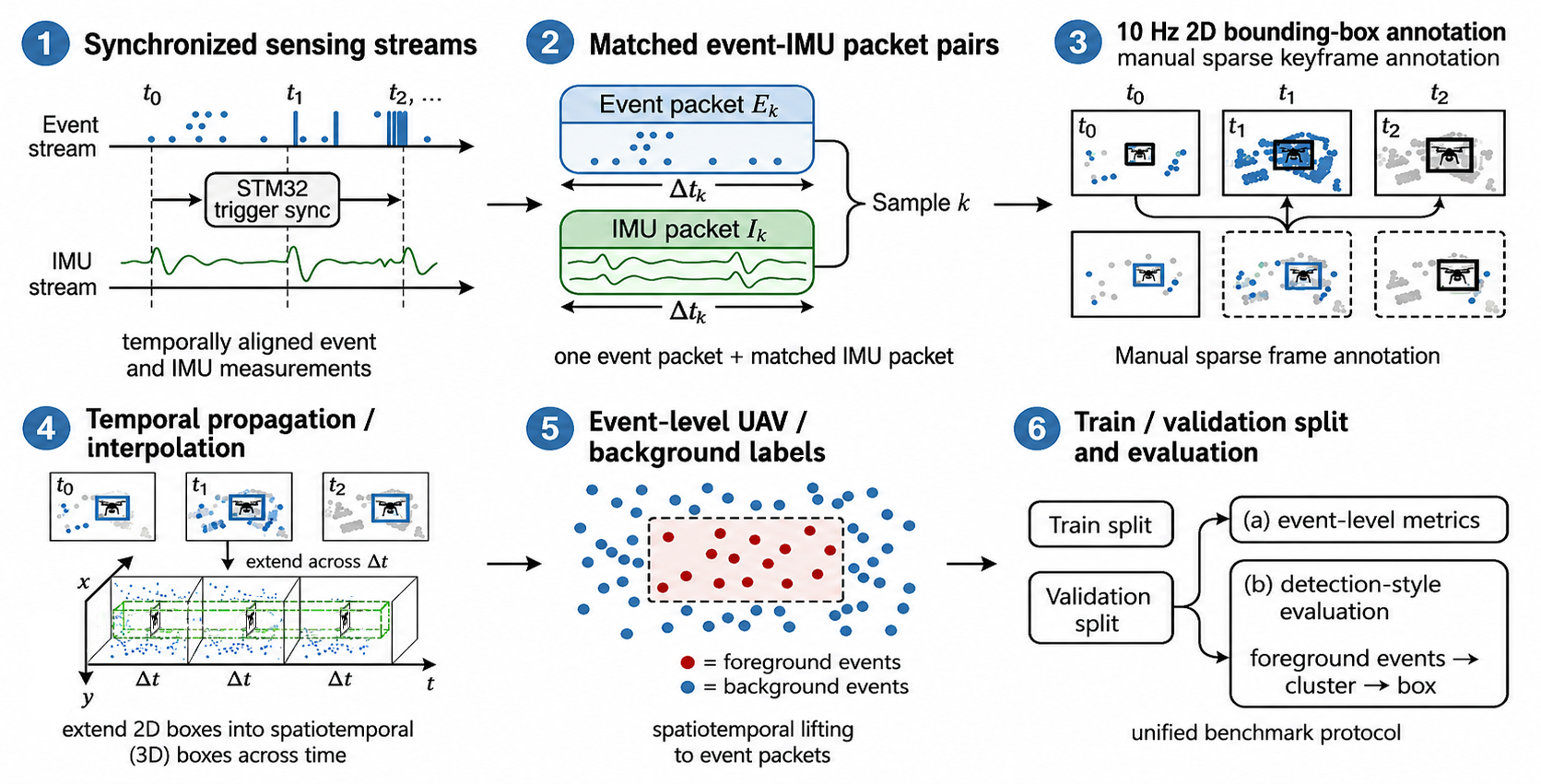}
  \end{tabular}
  \caption{
  \textbf{M\textsuperscript{2}E-UAV data collection and benchmark construction.} The top row shows the onboard sensing platform, including the carrier UAV, event camera, IMU, vibration-damping mount, and STM32-based trigger synchronization board. The bottom diagram summarizes how synchronized event and IMU streams are organized into matched event-IMU packets and paired with event-level UAV/background labels for benchmark evaluation. The packet construction and label lifting steps are detailed in Sections~\ref{sec:event_imu_packet_construction} and~\ref{sec:annotation_pipeline}.
  }
  \label{fig:benchmark_construction}
\end{figure}

\subsection{Event-IMU packet construction}
\label{sec:event_imu_packet_construction}
Each processed benchmark sample contains one event packet and its temporally matched IMU packet. The event packet is represented as
\begin{equation}
    \mathcal{E} = \{(x_i,y_i,t_i,p_i)\}_{i=1}^{N},
\end{equation}
where $(x_i,y_i)$ is the event location, $t_i$ is the timestamp inside the packet, and $p_i$ is polarity. Events remain asynchronous event data before being converted into supported representations, such as accumulated event frames, native event tensors, or event point sets.

The processed event packet stores event coordinates, timestamps, polarities, and binary UAV/background foreground labels. For point-set baselines, each sampled event is represented by $[x,y,t,p]$: image coordinates are normalized by the sensor resolution, timestamps are normalized within the packet, and polarity is mapped to $\{-1,1\}$. For point-set baselines only, we use a unified sampling protocol with 8,192 events per packet while retaining positive UAV events when available.

The matched IMU archive stores orientation, angular velocity, and linear acceleration. Each event packet in the train and validation splits has a corresponding IMU packet, which makes event-only and optional event-IMU settings directly comparable under the same sample index.

\subsection{Scene families and split}
\label{sec:scene_families_split}
Table~\ref{tab:scene_families} summarizes the train/validation packet counts across the four illumination-background regimes. Each sample is a matched event-IMU pair with 1280$\times$720 event resolution. The scenes cover different background structures and illumination conditions under onboard observer motion.

\begin{table}[t]
  \centering
  \caption{
  \textbf{Scene families.} Train and validation packets across the four illumination-background regimes.
  }
  \label{tab:scene_families}
  \renewcommand{\arraystretch}{0.9}
  \begin{tabular}{lcc}
    \toprule
    Scene family & Train packets & Val packets \\
    \midrule
    Sunny building-forest & 24,345 & 5,671 \\
    Sunny farm-village & 28,432 & 7,109 \\
    Sunset building-forest & 14,547 & 3,638 \\
    Sunset farm-village & 19,899 & 4,977 \\
    \bottomrule
  \end{tabular}
\end{table}

\begin{figure}[t]
  \centering
  \includegraphics[width = 1\textwidth]{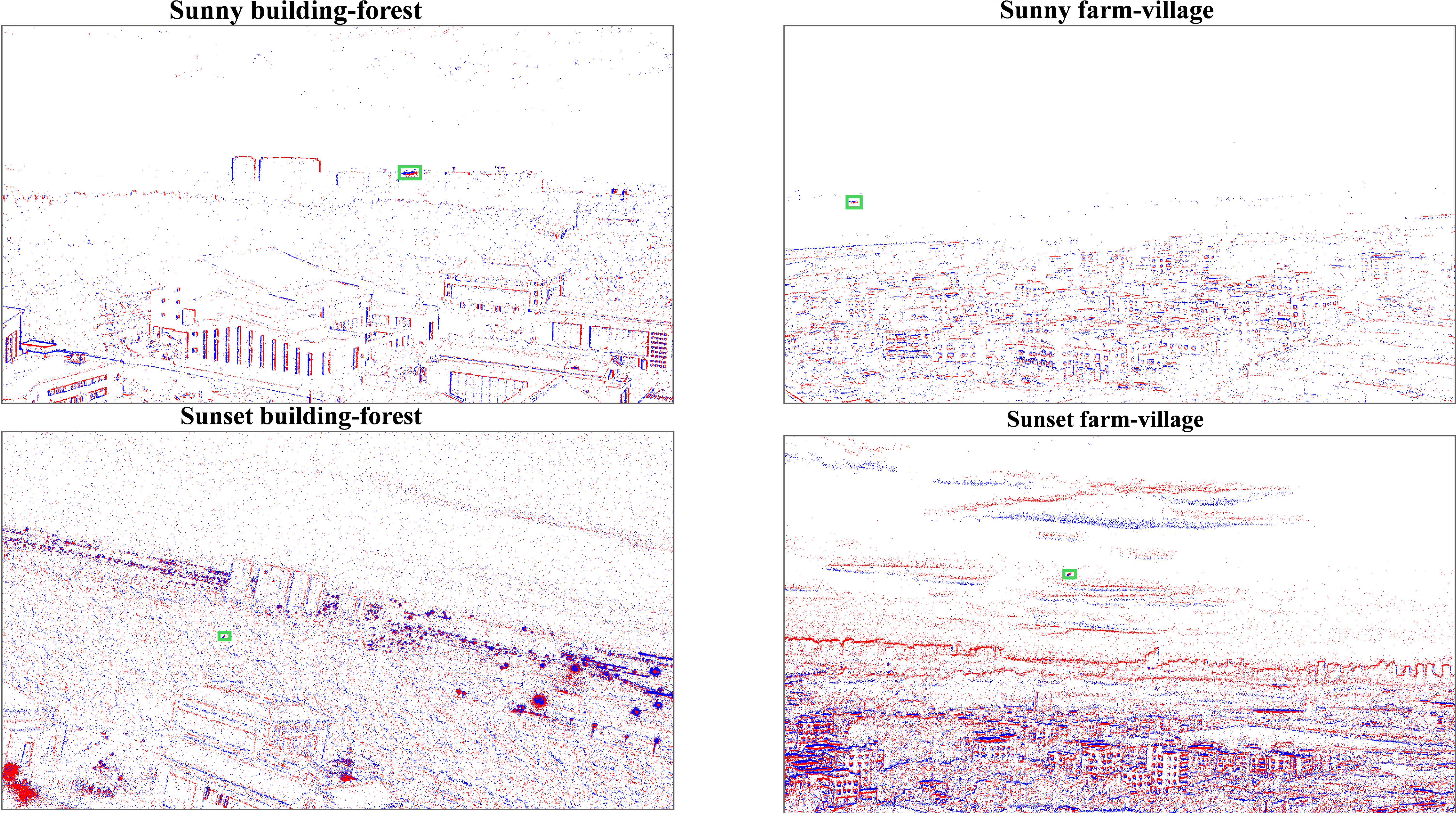}
  \caption{
  \textbf{Scene-family showcase.} The montage covers sunny/sunset illumination and building-forest/farm-village backgrounds.
  }
  \label{fig:scene_showcase}
\end{figure}

\subsection{Dataset statistics}
\label{sec:dataset_statistics}
The processed synchronized event-IMU files occupy 21.4 GB in total. The annotated UAV boxes have an average size of $22.91 \times 11.90$ pixels, corresponding to only $0.0296\%$ of the $1280 \times 720$ image area. This small spatial footprint confirms the tiny-target nature of M\textsuperscript{2}E-UAV and explains why sparse target events can be easily overwhelmed by ego-motion-induced background activity.

\subsection{Annotation pipeline}
\label{sec:annotation_pipeline}
We generate event-level supervision from sparse frame annotations. First, 2D UAV bounding boxes are annotated at 10 Hz. The sparse frame boxes are then propagated between adjacent labeled frames through temporal propagation and temporal interpolation, and this label propagation forms spatiotemporal supervision. % TODO(source): add a verified citation key for F-UAV-D-style sparse box propagation if available.
Similar to event-level label generation from box annotations in EV-UAV \citep{chen2025evuav}, the propagated spatiotemporal supervision is lifted to the corresponding event packets through spatiotemporal lifting. This event-level label generation produces point-level/event-level UAV foreground labels.

The resulting labels support both event-level foreground learning and box-level detection evaluation. The benchmark reports box-level metrics on the validation split while retaining event-level labels for methods that learn from sparse target evidence.

\subsection{Dataset comparison}
Table~\ref{tab:dataset_comparison} compares M\textsuperscript{2}E-UAV with representative event-based UAV, flying-object, and tracking datasets. The summarized properties are compiled from the benchmark and task definitions reported in the corresponding papers. The comparison focuses on properties that are central to onboard motion-on-motion tiny UAV detection: event resolution, UAV-centric collection, tiny-target setting, observer motion, synchronized IMU, and annotation granularity. Unlike datasets that provide only frame-level or bounding-box-level supervision, M\textsuperscript{2}E-UAV provides point-level/event-level UAV foreground labels derived from temporally propagated 10 Hz bounding-box annotations.

\begin{table}[H]
  \caption{
  \textbf{Comparison of event-based datasets along properties relevant to onboard tiny UAV detection.}
  }
  \label{tab:dataset_comparison}
  \centering
  \renewcommand{\arraystretch}{0.9}
  \resizebox{0.9\linewidth}{!}{%
  \begin{tabular}{lcccccc}
    \toprule
    Dataset & Resolution & UAV-centric & Tiny target & Observer motion & IMU & Annotation granularity \\
    \midrule
    EED~\citep{eed_dataset} & 240$\times$180 & \xmark & \xmark & \cmark & \xmark & frame-level \\
    VisEvent~\citep{visevent_dataset} & 346$\times$260 & \xmark & \xmark & \xmark & \xmark & frame-level \\
    EventVOT~\citep{eventvot_dataset} & 1280$\times$720 & \xmark & \xmark & \xmark & \xmark & frame-level \\
    F-UAV-D~\citep{fuavd_dataset} & 1280$\times$720 & \cmark & \xmark & \xmark & \xmark & frame-level \\
    EV-UAV~\citep{chen2025evuav} & 346$\times$240 & \cmark & \cmark & \xmark & \xmark & \textcolor{red}{point-level} \\
    EV-Flying~\citep{evflying_dataset} & 1280$\times$720 & \xmark & \xmark & \xmark & \xmark & frame-level \\
    NeRDD~\citep{nerdd_dataset} & 1280$\times$720 & \cmark & \xmark & \xmark & \xmark & frame-level \\
    CRSOT~\citep{crsot_dataset} & 1280$\times$720 & \cmark & \xmark & \cmark & \xmark & frame-level \\
    FRED~\citep{fred_dataset} & 1280$\times$720 & \cmark & \xmark & \xmark & \xmark & frame-level \\
    M\textsuperscript{2}E-UAV & 1280$\times$720 & \cmark & \cmark & \cmark & \cmark & \textcolor{red}{point-level} \\
    \bottomrule
  \end{tabular}%
  }
\end{table}

The comparison is not intended to reduce the contribution to individual dataset attributes; rather, it highlights that existing resources do not instantiate the same sensing problem, where a moving UAV-mounted event camera observes a moving tiny UAV target under strong ego-motion-induced background activity.

To the best of our knowledge, M\textsuperscript{2}E-UAV is the first event-based dataset and benchmark for onboard motion-on-motion tiny UAV detection, where a moving UAV-mounted event camera detects a moving tiny UAV target. This setting differs fundamentally from static- or ground-observer UAV detection because sensor ego-motion activates dense background events that can overwhelm the sparse target evidence.

\section{Task Definition and Evaluation Protocol}

\subsection{Task definition}
M\textsuperscript{2}E-UAV defines onboard motion-on-motion event-based tiny UAV detection. The input is an event packet
\begin{equation}
E = \{(x_i,y_i,t_i,p_i)\}_{i=1}^{N},
\end{equation}
optionally paired with synchronized IMU measurements from the same time interval. The expected output is a 2D UAV bounding box in the 1280$\times$720 event-camera image plane. Methods may use the event stream alone or use the synchronized IMU stream as an auxiliary onboard motion cue.

\subsection{Supported representations}
The benchmark is representation-agnostic. Event packets can be accumulated into event frames, discretized into voxel-grid or event-tensor representations, or kept as sparse point sets with coordinates, timestamps, and polarities. This allows event-frame detectors, event-tensor models, and point-set methods to be evaluated under the same train/validation split and annotation protocol.

\subsection{Evaluation protocol}
All benchmark results use the train/validation split in Section~\ref{sec:scene_families_split}. The validation split is used for reporting under a shared protocol, while held-out test evaluation is left to a separate benchmark release. The primary box-level metrics are mAP$_{50}$ and mAP$_{50:95}$. We also report detection-level F1 under the same evaluation protocol to summarize precision-recall trade-offs at a fixed operating point. For methods that output event-level foreground scores, event-level metrics can additionally be computed using the released foreground labels. This protocol is intended to characterize performance under sparse tiny-target evidence and dense ego-motion-induced background events without prescribing a particular network architecture.

\section{Experiments}

\subsection{Reporting protocol}
This submission uses the train/validation split described in Section~\ref{sec:scene_families_split} and Table~\ref{tab:scene_families}. Under this split, the validation set is used for validation-stage reporting, and held-out test evaluation is left to a separate benchmark release. The following table reports representative existing baselines under a unified protocol.

\subsection{Baseline implementation}
For each baseline, we follow the official implementation and default training configuration whenever available. Event streams are converted into the representation required by each baseline, including accumulated event frames, voxel grids, or point sets. All baselines are trained and evaluated on the same train/validation split.

\subsection{Benchmark comparison protocol}
Table~\ref{tab:benchmark_protocol} reports representative baselines evaluated under the unified validation protocol. The baseline set spans event-frame, event-tensor, and event-point representations: YOLOv10 on event images covers a dense image-style detector applied to event accumulations \citep{wang2024yolov10}; RVT and SAST cover recurrent and sparse event-detector families \citep{gehrig2023rvt,peng2024sast}; and KPConv, RandLA-Net, and COSeg cover point-set modeling directions \citep{thomas2019kpconv,hu2020randlanet,an2024coseg}. EV-UAV denotes the point-based baseline protocol/model released with the EV-UAV benchmark, rather than the EV-UAV dataset itself.

\begin{table}[H]
  \caption{
  \textbf{Benchmark results of representative baselines on M\textsuperscript{2}E-UAV.}
  }
  \label{tab:benchmark_protocol}
  \centering
  \renewcommand{\arraystretch}{0.9}
  \resizebox{0.7\linewidth}{!}{%
  \begin{tabular}{lcccc}
    \toprule
    Method & Representation & F1$\uparrow$ & mAP$_{50}$\(\uparrow\) & mAP$_{50:95}$\(\uparrow\) \\
    \midrule
    YOLOv10 & Event frame & 0.7283 & 0.6349 & 0.2566 \\
    RVT~\citep{gehrig2023rvt} & Voxel grid & 0.8643 & 0.8186 & 0.3815 \\
    SAST~\citep{peng2024sast} & Voxel grid & 0.6815 & 0.5519 & 0.3285 \\
    EV-UAV~\citep{chen2025evuav} & Point set & 0.0863 & 0.3761 & 0.1512 \\
    KPConv~\citep{thomas2019kpconv} & Point set & 0.4741 & 0.4289 & 0.2550 \\
    RandLA-Net~\citep{hu2020randlanet} & Point set & 0.3880 & 0.0228 & 0.0117 \\
    COSeg~\citep{an2024coseg} & Point set & 0.6254 & 0.5920 & 0.3124 \\
    \bottomrule
  \end{tabular}%
  }
\end{table}

The results show that existing baselines remain limited under sparse tiny-target evidence and dense ego-motion-induced background events. Event-frame and voxel-grid baselines provide useful coarse localization, but stricter mAP$_{50:95}$ remains low. Point-set transfers are also challenging, which highlights the need for methods designed around onboard motion-on-motion event data.

These results indicate three main challenges of M\textsuperscript{2}E-UAV. First, event-frame and voxel-grid methods can capture coarse motion structures but still suffer under strict localization metrics, as reflected by the gap between mAP$_{50}$ and mAP$_{50:95}$. Second, point-set baselines are sensitive to the severe foreground-background imbalance because tiny UAV events occupy only a small fraction of each packet. Third, ego-motion-induced background structures often generate dense events that compete with sparse target evidence. These observations suggest that M\textsuperscript{2}E-UAV is not only a dataset release but also a diagnostic benchmark for analyzing onboard motion-on-motion event perception.

\section{Conclusion}
We presented M$^2$E-UAV, to the best of our knowledge the first event-based dataset and benchmark for onboard motion-on-motion tiny UAV detection. Unlike static- or ground-observer settings, M$^2$E-UAV captures the case where a moving UAV-mounted event camera observes a moving tiny UAV target, producing dense ego-motion-induced background activity around sparse target evidence. The dataset, split, annotations, and benchmark protocol provide a reference setting for future event-only and event-IMU comparisons. The benchmark results show that existing event-frame, event-tensor, and event-point baselines still struggle with sparse tiny-target evidence and dense ego-motion-induced background events, indicating that onboard motion-on-motion event-based tiny UAV detection remains an open challenge.

\textbf{Limitations.}
M\textsuperscript{2}E-UAV currently focuses on four representative illumination-background regimes and uses a validation-stage protocol for onboard motion-on-motion event-based tiny UAV detection. Future releases can further expand the benchmark with more weather, altitude, distance, and target-motion conditions. 
{
\small
\bibliographystyle{plainnat}
\bibliography{refs}
}

\end{document}